\documentclass{article} 
\usepackage{iclr2026_conference,times}
\iclrfinalcopy

\usepackage{amsmath,amsfonts,bm}

\def\eqref#1{equation~\ref{#1}}

\def\1{\bm{1}}

\DeclareMathAlphabet{\mathsfit}{\encodingdefault}{\sfdefault}{m}{sl}
\SetMathAlphabet{\mathsfit}{bold}{\encodingdefault}{\sfdefault}{bx}{n}

\usepackage{hyperref}
\usepackage{url}
\usepackage{booktabs} 
\usepackage{multirow} 
\usepackage{caption} 
\usepackage{graphicx}
\usepackage{tabularx}
\usepackage{adjustbox}
\usepackage{booktabs} 
\usepackage{graphicx}
\usepackage{subcaption}
\usepackage{amsmath}
\usepackage{amsfonts}
\usepackage{booktabs}
\usepackage{hyperref}
\usepackage{enumitem}
\usepackage{wrapfig}
\usepackage{algorithm}
\usepackage{algorithmic}
\usepackage{subcaption}
\usepackage{xcolor}
\usepackage{cleveref}

\usepackage[T1]{fontenc}
\usepackage{times}
\usepackage{changes}

\title{TimeSense: Making Large Language Models Proficient in Time‑Series Analysis}

\author{
Zhirui Zhang\textsuperscript{1} \quad
Changhua Pei\textsuperscript{2} \quad
Tianyi Gao\textsuperscript{3} \quad
Zhe Xie\textsuperscript{1} \quad
Yibo Hao\textsuperscript{1} \\
{ \textbf{
Zhaoyang Yu\textsuperscript{1} \quad
Longlong Xu\textsuperscript{1} \quad
Tong Xiao\textsuperscript{1} \quad
Jing Han\textsuperscript{3} \quad
Dan Pei\textsuperscript{1} }}
\\
\textsuperscript{1}Tsinghua University \\
\textsuperscript{2}Computer Network Information Center, Chinese Academy of Sciences \\
\textsuperscript{3}ZTE Corporation
\\
\texttt{zr-zhang25@mails.tsinghua.edu.cn}
}

\newcommand{\tsbench}{EvalTS}

\begin{document}
\renewcommand{\sectionautorefname}{$\S$}
\renewcommand{\subsectionautorefname}{$\S$}
\renewcommand{\subsubsectionautorefname}{$\S$}

\maketitle
\begin{figure}[h]
    \centering
    \includegraphics[width=1.\linewidth]{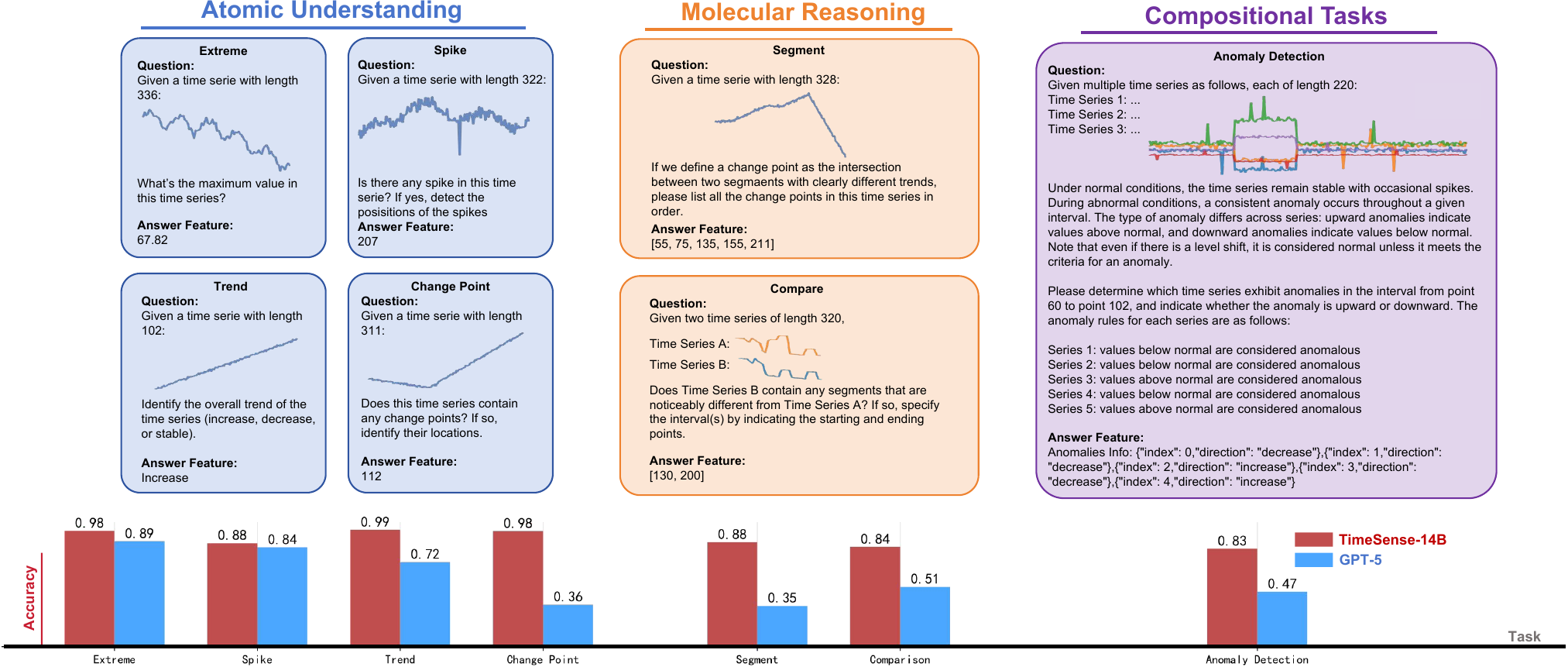}
    \caption{Examples from the EvalTS benchmark and performance comparison between our model and GPT-5.}
    \label{fig:qa_example}
\end{figure}
\begin{abstract}
\vspace{-10pt}
In the time-series domain, an increasing number of works combine text with temporal data to leverage the reasoning capabilities of large language models(LLMs) for various downstream time-series understanding tasks. This enables a single model to flexibly perform tasks that previously required specialized models for each domain. However, these methods typically rely on text labels for supervision during training, biasing the model toward textual cues while potentially neglecting the full temporal features. Such a bias can lead to outputs that contradict the underlying time-series context. To address this issue, firstly, we construct the EvalTS benchmark, comprising 10 tasks across three difficulty levels, from fundamental temporal pattern recognition to complex real-world reasoning, to evaluate models under more challenging and realistic scenarios. We also propose TimeSense, a multimodal framework that makes LLMs proficient in time-series analysis by balancing textual reasoning with a preserved temporal sense. TimeSense incorporates a Temporal Sense module that reconstructs the input time-series within the model’s context, ensuring that textual reasoning is grounded in the time-series dynamics. Moreover, to enhance spatial understanding of time-series data, we explicitly incorporate coordinate-based positional embeddings, which provide each time point with spatial context and enable the model to capture structural dependencies more effectively. Experimental results demonstrate that TimeSense achieves state-of-the-art performance across multiple tasks, and it particularly outperforms existing methods on complex multi-dimensional time-series reasoning tasks. 
\end{abstract}

\section{introduction}
\label{sec:intro}
Time series data lie at the core of domains such as electricity, healthcare, traffic, weather, and finance~\citep{nogales2002forecasting,morid2023time,lippi2013short,mcgovern2011identifying,yu2023temporal}. Recent advances in multimodal learning show that combining language with temporal signals can unlock stronger reasoning, echoing the progress seen in vision-language modeling.~\citep{jin2023unified,wang2023visionllm} Time-series multimodal models extend this idea by using large-scale pretraining to perform diverse tasks in a zero-shot setting.~\citep{xie2024chatts,wang2024chattime,xu2025itformer,kong2025time} 

Yet, in many of these models, text serves only as an auxiliary signal to boost performance on predefined temporal tasks, rather than enabling flexible adaptation through context~\citep{jin2023time,yang2025timerag}. Natural language, however, is more than structured labels. It offers rich descriptions of temporal patterns and a human-interpretable reasoning process. This motivates the design of multimodal time-series models that not only solve tasks but also explain them in natural language, leading to richer reasoning and more intuitive interaction.

Despite this promise, current approaches face two key challenges:

(\romannumeral 1) Datasets remain narrow. Classical time series tasks focus on low-level tasks such as forecasting or classification. While useful, these tasks capture only basic dynamics and cannot test whether a model generalizes to reasoning that links temporal and textual information. Existing reasoning datasets emphasize surface-level alignment between text and time series, but do not test deeper temporal sense or cross-feature reasoning.

(\romannumeral 2) Most models are trained with only textual labels, which biases optimization toward language and weakens temporal modeling. This often produces outputs that contradict the underlying sequence, especially for long series, multi-dimensional dynamics, or local anomalies. A more balanced method is needed—one that preserves temporal sense while still using textual supervision.

To address challenge of datasets, we introduce EvalTS, a benchmark designed to comprehensively assess multimodal time-series models in a dialog-style manner, capturing both their perceptual and reasoning abilities. EvalTS draws inspiration from human temporal understanding, rather than being constrained to fixed tasks or simple alignment-based evaluations common in existing benchmarks, starting from atomic units of temporal cognition and progressively combining them to tackle more complex tasks. To support this evaluation, we propose ChronGen, a controllable rule-based generator that systematically creates multimodal time-series data, filling the gap of data scarcity. 

To enable models to better tackle such complex tasks, we propose TimeSense, an architecture that mimics human temporal analysis by encoding each time point’s positional information and integrating a Time Sensor module, allowing the model to retain, reason over, and fully exploit temporal information. Together, these contributions enable models to achieve holistic temporal awareness and effectively leverage both temporal and contextual cues for complex time-series tasks.

In summary, our contributions are threefold:

\textbf{1.} We design EvalTS, a benchmark for systematic evaluation across diverse temporal reasoning tasks, going beyond simple alignment tasks and classical temporal tasks to capture a broader spectrum of multimodal time-series analysis challenges.

\textbf{2.} We propose TimeSense, which integrates temporal and textual information through a Temporal Sense module, enabling the model to process contextual text and temporal signals in a balanced, human-like manner and mitigating language bias in multimodal training.

\textbf{3.} We show that TimeSense consistently outperforms existing models across all 10 tasks in EvalTS and remains highly competitive on four additional multiple-choice evaluation datasets, demonstrating its robust and generalizable time-series analysis capabilities.

\section{Related Works}
\label{sec:relatedworks}
\subsection{Classical Time-Series Tasks and Methods}
Research on time series has centered on forecasting, classification, clustering, change-point detection, anomaly detection, and causal inference~\cite{zhou2021informer,dempster2020rocket,tiwari2023segmentation,gong2024causal}. Early methods used statistical models such as ARIMA, exponential smoothing, and state-space models~\citep{zhang2003time,kalekar2004time}, which explicitly encode temporal dependencies. Later, deep learning models such as RNNs, LSTMs, and temporal convolutional networks~\citep{liu2020dstp,siami2018comparison,he2019temporal} showed stronger ability to capture non-linear dynamics.

Many of these methods address task-specific temporal patterns. Pattern-based approaches have been common in classification and motif discovery, where recognizing local temporal structures improves accuracy~\citep{ye2009time,zakaria2012clustering}. Similarly, anomaly detection and change-point detection often depend on identifying recurring or distinctive temporal signatures. Overall, classical methods focus on patterns in temporal data itself, with little use of textual or multimodal signals.

\subsection{Time-Series Multimodal Models}

In recent years, multimodal large language models (MLLMs) have achieved rapid progress in images, videos, and audio~\citep{zhang2005probabilistic,wang2024internvideo2,mroueh2015deep}. These models use natural language understanding to support reasoning, question answering, and decision making.

By contrast, multimodal modeling for time series is still in its early stages. Work in this area is limited by the lack of datasets that pair sequences with text. Emerging time-series MLLMs attempt to connect numerical signals with language, often producing textual rationales or answers. For example, ChatTime~\cite{wang2024chattime} treats time series as a “foreign language,” enabling bi-modal inputs and zero-shot forecasting. ChatTS~\cite{chow2024chatts} aligns time-series attributes with LLMs using synthetic descriptions to improve reasoning. ITFormer links a time-series encoder with a frozen LLM and introduces a domain-specific multi-task benchmark, EngineMT-QA, to evaluate temporal-textual reasoning~\citep{xu2025itformer,engineMTQA2025}. Survey papers summarize this fast-moving field and categorize approaches by encoders, prompting strategies, and training objectives~\citep{zhu2024surveytsllm}.

\section{Methodology}
\label{sec:method}

\subsection{Problem Definition}
\begin{figure}[t]
    \centering
    \includegraphics[width=0.9\linewidth]{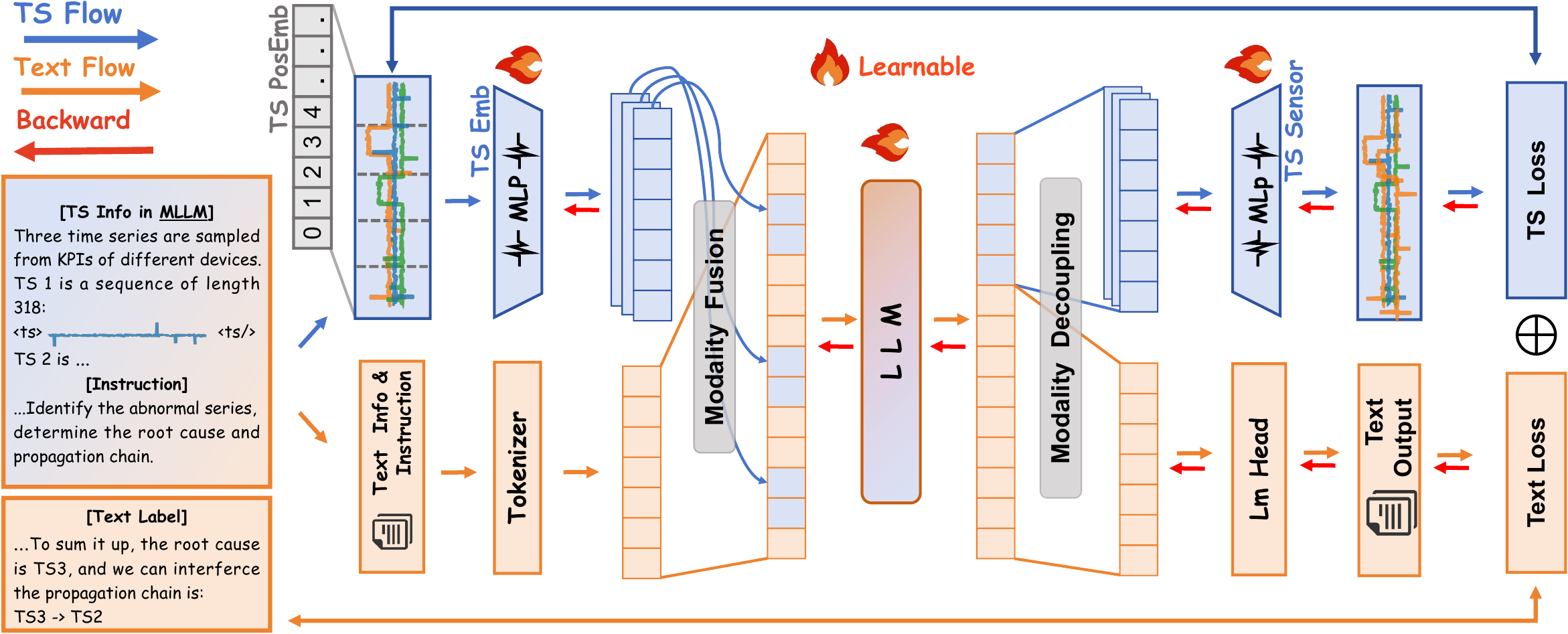}
    \caption{\textbf{Workflow of TimeSense.} 
    The time-series related modules are marked with blue lines, and the text-related modules with orange lines.}
    \label{fig:pipeline}
\end{figure}

We first present the motivation and the challenges that TimeSense aims to solve, followed by the design details. As shown in~\autoref{fig:pipeline}, the multi-modal architecture consists of two main modules: modality embedding and modality sensor. After receiving multimodal inputs, the textual and temporal modalities are processed along separate flows. The time series is transformed within the time series(TS) flow (blue path) into TS tokens that can be embedded into a language model. These tokens are aligned with tokenized text according to their original positions. The mixed tokens are then fed into an LLM for reasoning.

At the output stage, the tokens are split into TS tokens and text tokens. The temporal component is reconstructed into a multivariate time series, which serves as a supervision signal to ensure faithful modeling of temporal dynamics. The textual component is decoded into natural language via a tokenizer, producing human-interpretable reasoning outputs for the target tasks~\citep{wei2022cot, wang2022selfconsistency, openai2023gpt4, touvron2023llama2}.

The goal of a Time Series Multimodal Large Language Model (TS-MLLM) is to jointly leverage multivariate time series and contextual text for question answering and reasoning. Formally, the multivariate time series is 
\[
\mathbf{X} = \{x_0, x_1, \ldots, x_{L-1}\} \in \mathbb{R}^{D \times L},
\]
where $D$ is the dimensionality and $L$ the sequence length. The textual information is a natural language instruction $\mathbf{I}$, which specifies contextual knowledge and the task. We abstract the TS-MLLM as a mapping function $\phi$, and the task can be formulated as $A = \phi(\mathbf{I}, \mathbf{X}),$ where $A$ is the predicted answer.

\subsection{Time Series Embedding}
\label{subsec:timeseriesembedding}
\begin{wrapfigure}{r}{0.48\textwidth} %
  \centering
  \vspace{-10pt} 
  \begin{subfigure}{0.48\linewidth}
      \centering
      \includegraphics[width=\linewidth]{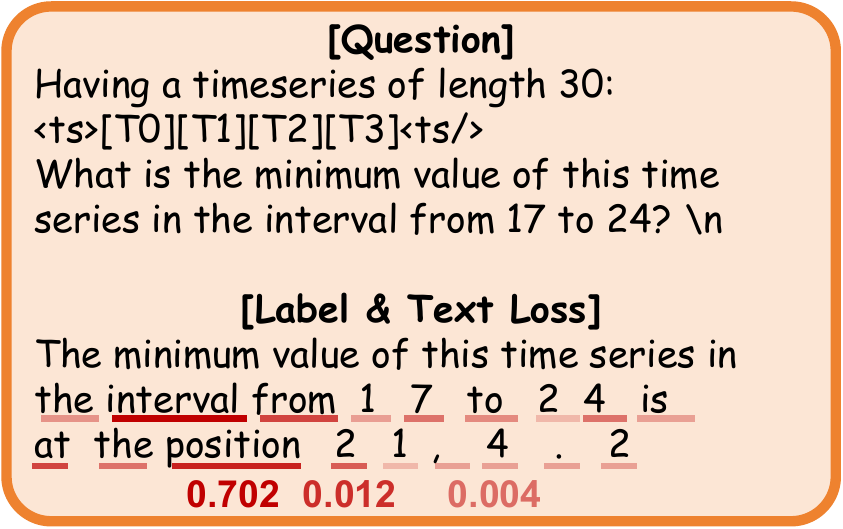}
      \caption{Loss distribution.}
      \label{fig:sense_loss}
  \end{subfigure}
  \hfill
  \begin{subfigure}{0.48\linewidth}
      \centering
      \includegraphics[width=\linewidth]{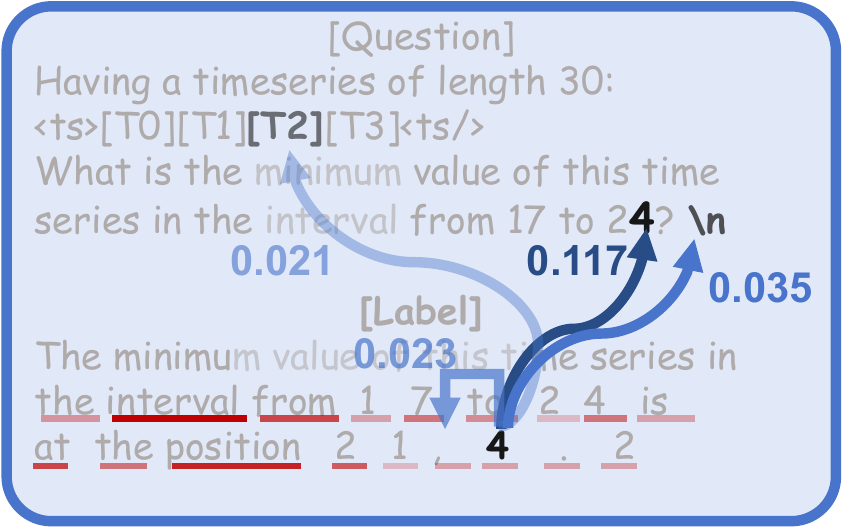}
      \caption{Attention distribution.}
      \label{fig:sense_attention}
  \end{subfigure}
  \caption{Token-level {loss and attention} of label tokens, showing how temporal information is overshadowed by textual content during model optimization.}
  \label{fig:attn_loss}
  \vspace{-5pt} 
\end{wrapfigure}
\textbf{Motivation.}  
A central challenge in multimodal large models lies in how to transform and integrate heterogeneous modalities into a unified token space. For time-series signals, prior work~\cite{nie2023patchtst} widely adopts a patching strategy, which segments sequences into local chunks to reduce length and capture short-range dependencies. While efficient, this compression introduces two challenges: (\romannumeral1) absolute temporal positions are lost once consecutive points are merged;(\romannumeral2) patching only preserves local structures. As illustrated in~\autoref{fig:wrap_ablation2}, neglecting absolute indices significantly degrades temporal reconstruction and reasoning.

\textbf{Solution.}  
We design a temporal embedding pipeline to address the above challenges. First, to retain absolute positions, we explicitly inject the index of each time step as an additional dimension alongside its value. Second, to preserve local relative dependencies while ensuring computational efficiency, we follow the patching strategy of~\cite{xie2024chatts, nie2023patchtst} with non-overlapping patches. This fusion ensures that each token encodes not only local temporal context but also the absolute coordinate of the underlying segment, providing the model with richer information for downstream tasks.

\textbf{Details.}  
Let the input be $\mathbf{X}=\{x_t^{(m)} \mid t=0,\dots,L-1;\ m=1,\dots,D\}\in\mathbb{R}^{L\times D}$, where $D$ is the number of channels and $L$ is the sequence length. In the following, we describe how the raw time series is transformed through positional encoding and tokenization, and eventually fused with other modalities.

A) \textit{Time Series Position Embedding.}  
We process each channel independently. For channel $d$, the sequence is $\mathbf{x}^{(d)}=[x_0^{(d)},\dots,x_{L-1}^{(d)}]\in\mathbb{R}^{L}$, and the absolute index vector is $\mathrm{Index}=[0,1,\dots,L-1]\in\mathbb{R}^{L}$. We then augment values with indices by concatenation: $\tilde{\mathbf{X}}^{(d)}=\big[\mathrm{Index};\mathbf{x}^{(d)}\big]\in\mathbb{R}^{2\times L}$. Stacking across channels yields $\tilde{\mathbf{X}}\in\mathbb{R}^{D\times 2\times L}$, which retains both absolute positions and values.

B) \textit{Patching.}  
Each augmented channel is split into non-overlapping patches of length $P$. For a sequence of length $L$, the number of patches is $N^{(L)}=\lceil L/P \rceil$, where zero-padding is applied to the last patch if needed. For channel $d$ and patch index $n \in \{0,\dots,N^{(L)}-1\}$, the flattened segment is defined as $\hat{\mathbf{x}}^{(d)}_{n} = \operatorname{reshape}(\tilde{\mathbf{X}}^{(d)}_{:,\,nP:(n+1)P-1}) \in \mathbb{R}^{2P}$. Collecting all patches yields $\hat{\mathbf{X}}^{(L)} \in \mathbb{R}^{D \times N^{(L)} \times 2P}$. When constructing a batch containing multiple sequences with different lengths $\{L_i\}$, the corresponding patch numbers $\{N^{(L_i)}\}$ may vary. To enable efficient batch training, we set $N=\max_i N^{(L_i)}$ and apply zero-padding along the patch dimension for all sequences with $N^{(L_i)} < N$. Thus, the final embedded representation for the whole batch is consistently aligned as $\hat{\mathbf{X}} \in \mathbb{R}^{B \times D \times N \times 2P}$, where $B$ is the batch size.

C) \textit{MLP Encoder.}  
To map each patch into the hidden dimension $H$, we apply a shared MLP $f_{\phi}:\mathbb{R}^{2P}\to\mathbb{R}^{H}$. Each patch token is $\mathbf{T}_{ts}^{(d,n)}=f_{\phi}(\hat{\mathbf{x}}^{(d)}_n)\in\mathbb{R}^{H}$, forming $\mathbf{T}_{ts}\in\mathbb{R}^{D\times N\times H}$. For transformer-style consumption, we flatten channel and patch axes to $\mathbf{T}_{ts}^{\mathrm{flat}}\in\mathbb{R}^{(D\cdot N)\times H}$, producing flat tokens suitable for modality joint with text tokens.

D) \textit{Modality Fusion.}  
For each extracted time-series segment $\hat{\mathbf{X}}$, the MLP encoder yields flat tokens
\[
\mathbf{T}_{ts}^{\mathrm{flat}} 
= \big[ \mathbf{T}_{ts}^{(1)}, \mathbf{T}_{ts}^{(2)}, \dots, \mathbf{T}_{ts}^{(M)} \big] 
\in \mathbb{R}^{(D \cdot N)\times H},
\]
where each sub-sequence $\mathbf{T}_{ts}^{(m)} \in \mathbb{R}^{N_m \times H}$ corresponds to the $m$-th original series segment, and $N_m$ matches the number of patches derived from that segment.  
Preserving temporal order, these sub-sequences are inserted into the textual sequence at designated positions, enclosed by special markers \texttt{<ts>} and \texttt{<ts/>}.  
This design ensures seamless alignment between temporal and textual tokens for joint modeling. 

\subsection{Time Series Sensing}
\textbf{Motivation.}  
A common paradigm in multimodal reasoning is to encode non-text modalities (e.g., time series) and feed them into an LLM, while using text as the sole output channel. This design has a clear advantage: it naturally leverages the strong reasoning and generation capabilities of LLMs, enabling seamless interaction with humans in natural language. However, this approach also comes with a critical drawback. Since the training objective is dominated by text-based supervision, the model tends to gradually ignore the non-text modalities. As illustrated in \autoref{fig:attn_loss}, text-related components in the label contribute significantly more to the loss than numerical tokens that reflect temporal information. Similarly, the attention visualization in \autoref{fig:sense_attention} shows that when generating time-series-related tokens, the model primarily attends to the textual instructions rather than temporal features. Consequently, the model learns only shallow text matching and fails to capture the structural patterns of time series. In short, relying solely on text labels prevents the model from forming a holistic sense of multi-modal time series.  

\textbf{Solution.}  
\begin{wrapfigure}{r}{0.48\textwidth}
    \centering
    \vspace{-10pt} 
    \begin{minipage}{\linewidth}
        \centering
        \begin{subfigure}[t]{0.32\linewidth}
            \includegraphics[width=\linewidth]{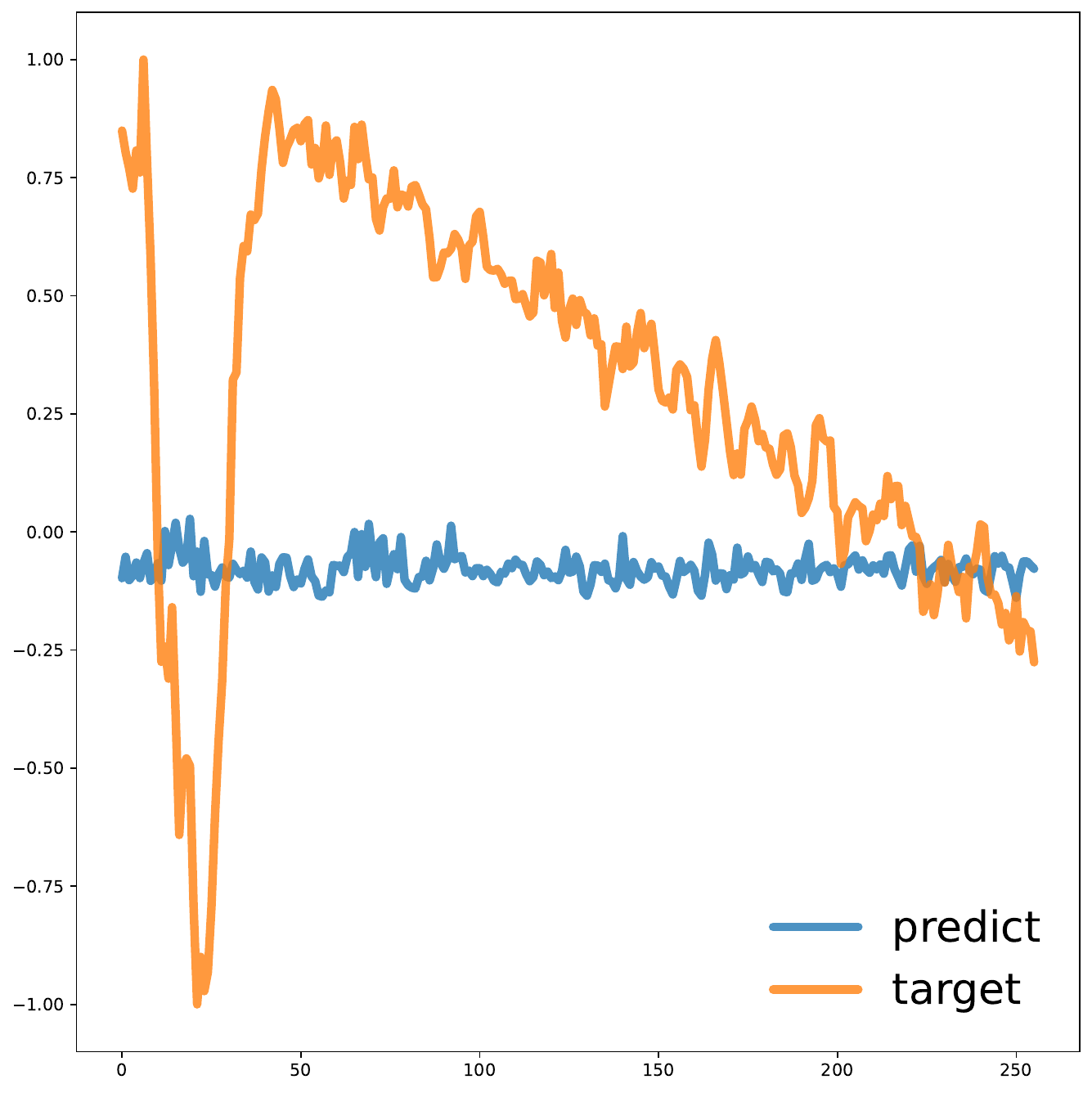}
            \caption{w/o FFT loss}
            \label{fig:wrap_ablation1}
        \end{subfigure}%
        \hfill
        \begin{subfigure}[t]{0.32\linewidth}
            \includegraphics[width=\linewidth]{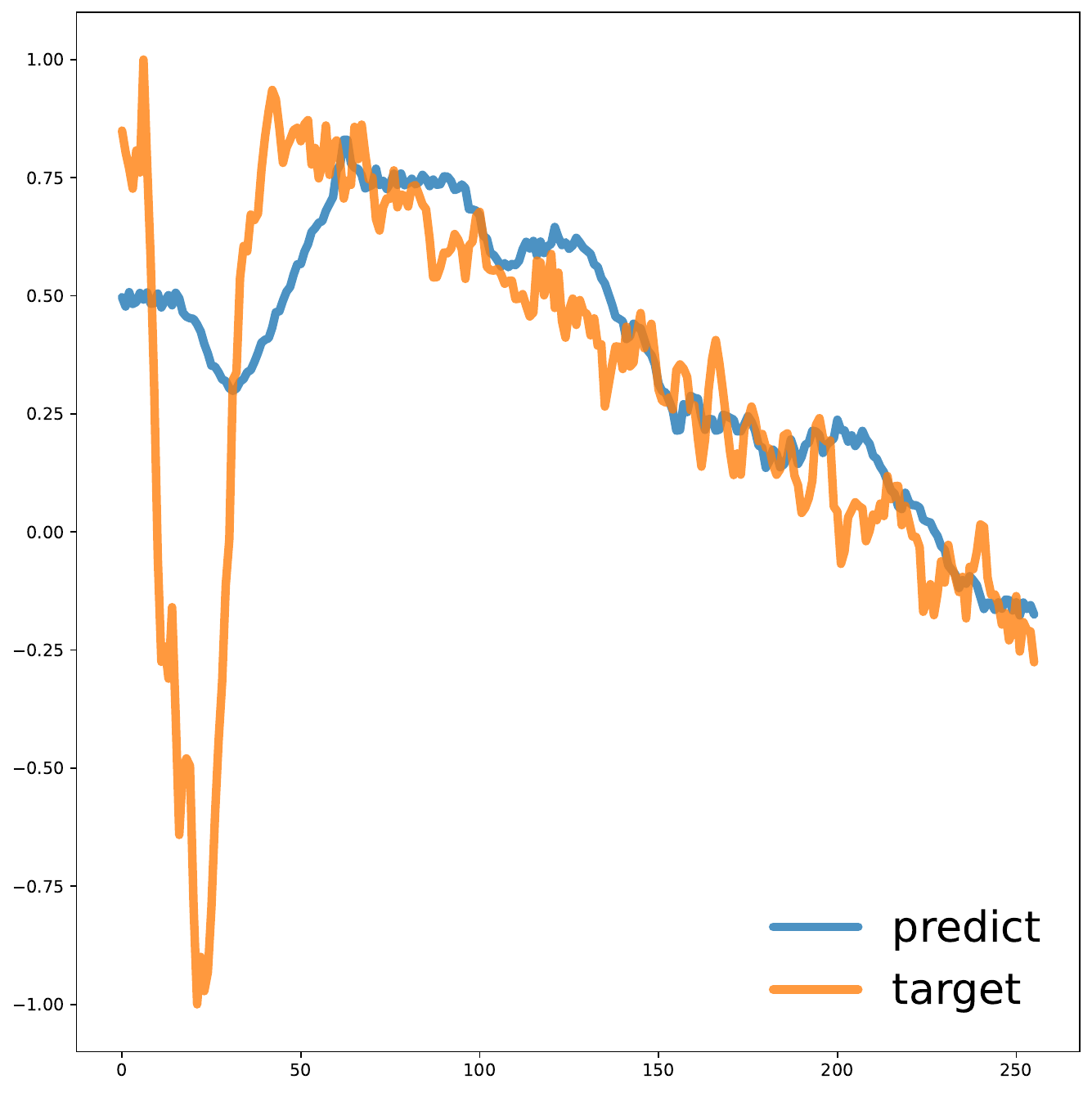}
            \caption{w/o PosEmb}
            \label{fig:wrap_ablation2}
        \end{subfigure}%
        \hfill
        \begin{subfigure}[t]{0.32\linewidth}
            \includegraphics[width=\linewidth]{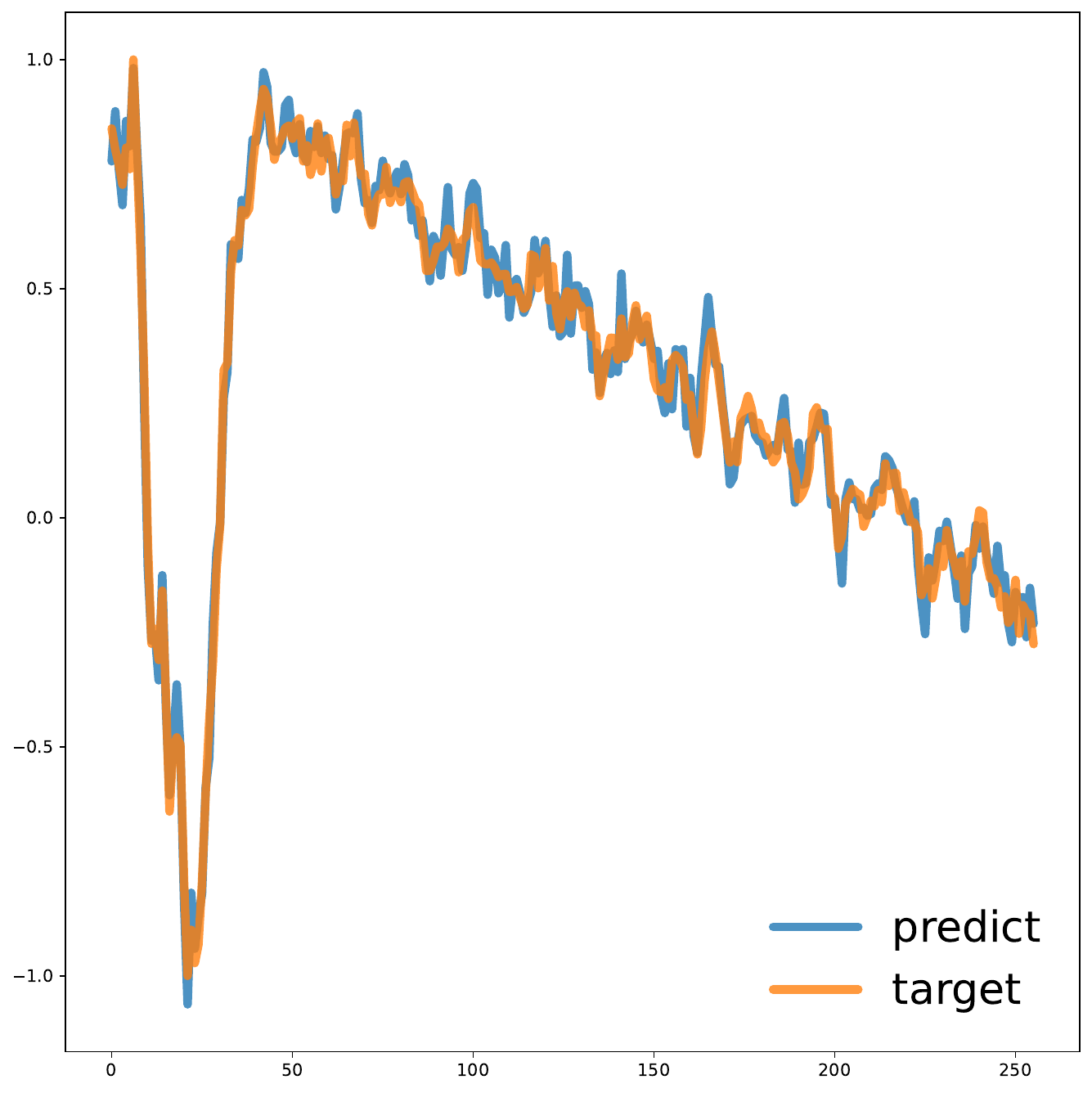}
            \caption{Full Conf}
            \label{fig:wrap_ablation3}
        \end{subfigure}
    \end{minipage}
    \caption{Temporal reconstruction results under different configurations.}
    \label{fig:wrap_ablation}
    \vspace{-15pt} 
\end{wrapfigure}
The key challenge is the lack of explicit supervision for the time-series modality. To address this, we require the model to decode its hidden time-series tokens into multi-dimensional sequences that reconstruct the original input, ensuring that temporal information is explicitly preserved during training.

\textbf{Details.}  
Similar to standard LLMs reports ~\cite{bai2025qwen2,touvron2023llama}, the model generates a hidden representation $\mathbf{h}_t \in \mathbb{R}^d$ for each token. We extend this process by introducing the following time-series-aware components:

\textit{A) Modality Decoupling.}  
The hidden output $\mathbf{H}=\{\mathbf{h}_1,\dots,\mathbf{h}_T\}$ contains mixed information from both modalities. To avoid interference, we designate the first $D\cdot N$ tokens as time series tokens $\mathbf{H}^{(ts)}=\{\mathbf{h}_1,\dots,\mathbf{h}_{D\cdot N}\}$, while the remaining tokens $\mathbf{H}^{(txt)}=\{\mathbf{h}_{{D\cdot N}+1},\dots,\mathbf{h}_T\}$ are used for conventional text modeling.  

\textit{B) Time-Series Reconstruction.}  
The latent representation is $\mathbf{H}^{(ts)} \in \mathbb{R}^{(D \cdot N) \times H}$, which can be viewed as $D \cdot N$ tokens of dimension $H$. Applying the MLP decoding $f_{\phi_\text{r}}: \mathbb{R}^{H} \to \mathbb{R}^{P}$ independently to each token yields $\hat{\mathbf{X}}_{\text{rec}}^{(d)} \in \mathbb{R}^{(D \cdot N) \times P}$. Re-arranging this tensor into channel token form gives $\hat{\mathbf{X}}_{\text{rec}}^{(d)} \in \mathbb{R}^{D \times N \times P}$, and finally the inverse patching merges $(N,P)$ into the original sequence length $L$, producing $\mathbf{X}_{\text{rec}} \in \mathbb{R}^{D \times L}$. Formally, the above process can be expressed as follows:
\[
(D \cdot N) \times H 
\;\xrightarrow{f_{\phi_r}}\; (D \cdot N) \times P
\;\xrightarrow{\text{reshape}}\; D \times N \times P
\;\xrightarrow{\text{inv-patch}}\; D \times L
\]

\textit{C) Time-Series Loss.}  
Unlike most time series foundation models trained solely on time-series targets (~\autoref{fig:wrap_ablation1}), using only MSE is insufficient when optimizing both time-series and textual objectives. Small point-wise variations contribute little to the MSE, especially compared to the textual cross-entropy loss, making it difficult for the time-series module to capture temporal patterns. To address this, we introduce a frequency-domain loss that captures high-frequency variations, following~\cite{wang2024fredf}, which shows that frequency-domain targets can enhance model performance. To preserve both value fidelity and temporal dynamics, we combine time-domain and frequency-domain constraints.
\[
\mathcal{L}_{ts} = \|{\mathbf{X_{rec}}} - \mathbf{X}\|_2^2 + \|\mathcal{F}(\mathbf{{X_{rec}}}) - \mathcal{F}({\mathbf{X}})\|_2^2,
\]  
where $\mathcal{F}(\mathbf{X})=\sum_{t=0}^{L-1}x_t e^{-j 2\pi kt / L}$ denotes the discrete Fourier transform (DFT).  

\textit{D) Loss Integration and Training.}  
For the textual part, we only consider the hidden states
corresponding to the text tokens, i.e., $\mathbf{H}^{(\text{txt})} = \{\mathbf{h}_{{D\cdot N}+1}, \dots, \mathbf{h}_T\}.$
Each hidden state $\mathbf{h}_i$ is projected into the vocabulary space
by the language modeling head: 
$\mathbf{z}_i = lm\_head(\mathbf{h}_i) \quad i=D\cdot N+1,\dots,T.$
where $\mathbf{z}_i \in \mathbb{R}^{|\mathcal{V}|}$ denotes the logits.

The predicted distribution is obtained via softmax: $\hat{\mathbf{y}}_i = \text{softmax}(\mathbf{z}_i) \in [0,1]^{|\mathcal{V}|}$. 
Finally, the textual cross-entropy loss is computed as
$\mathcal{L}_{\text{txt}} 
= - \frac{1}{T-D\cdot N} \sum_{i=D\cdot N+1}^{T} 
\mathbf{y}_i^\top \log \hat{\mathbf{y}}_i ,$
where $\mathbf{y}_i \in \{0,1\}^{|\mathcal{V}|}$ is the one-hot label for
the $i$-th token.

The final training objective combines both constraints: 
$\mathcal{L} = \mathcal{L}_{\text{txt}} + \mathcal{L}_{\text{ts}}$,  
jointly updating the LLM backbone, the time-series embedding, and the reconstruction MLP during back propagation.

\section{ChronGen}
\label{sec:ChronGen}
To train and evaluate multimodal time-series models on their ability to recognize and utilize various temporal features, we propose ChronGen, a data generator that incrementally incorporates local features into a base trend line and annotates these features using natural language. ChronGen produces comprehensive representations of time series, from which we construct question–answer pairs by selecting and combining local features, thereby enabling the training of models in temporal cognition and reasoning. To thoroughly assess this capability, we design three difficulty levels covering a total of eleven types of temporal reasoning tasks, as detailed in \autoref{subsec:tsbench}.

\subsection{Implementation}
\label{subsec:chrongen_implementation}
\begin{wrapfigure}{r}{0.48\textwidth} %
  \centering
  \vspace{-10pt}
  \includegraphics[width=\linewidth]{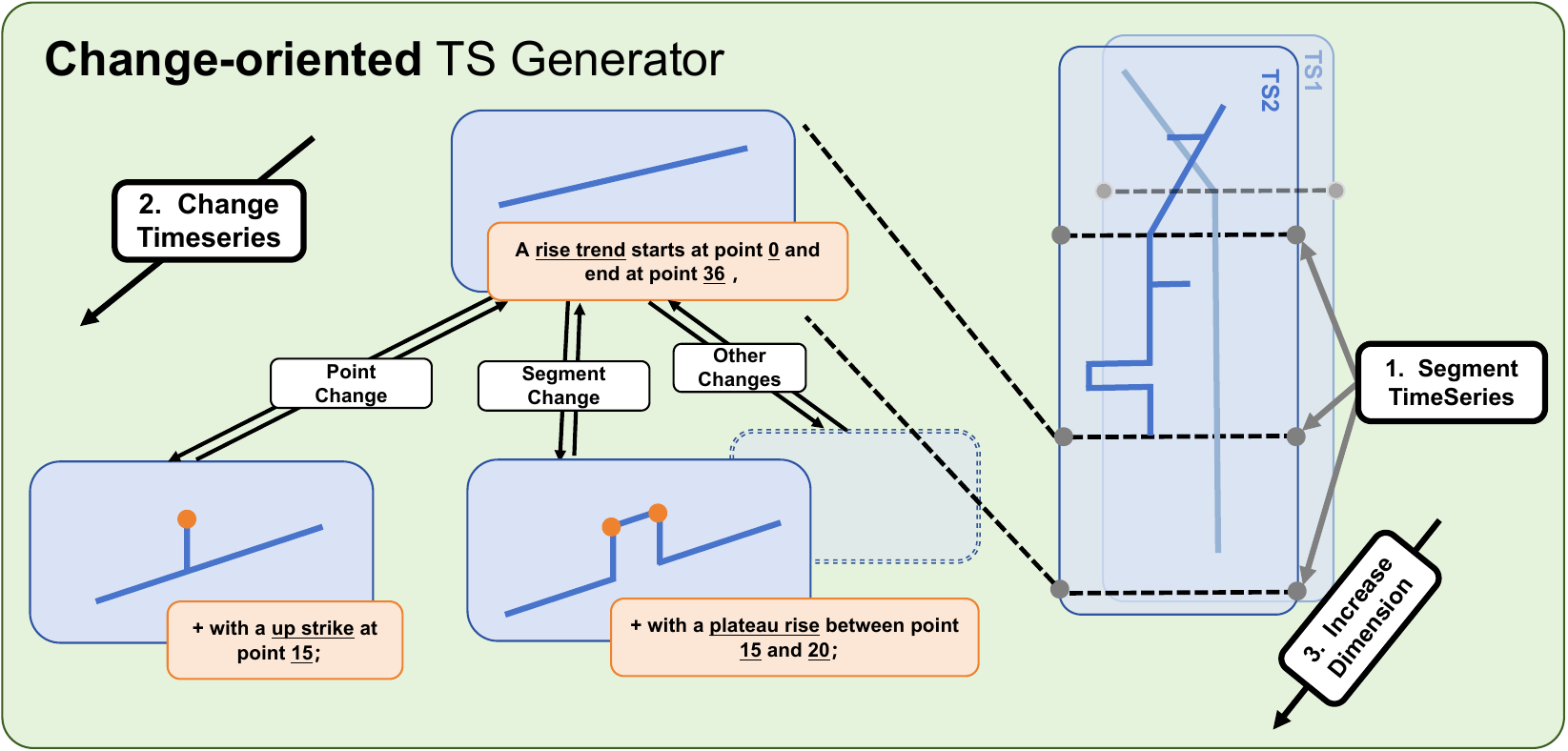}
  \hfill
  \vspace{-15pt}
  \caption{Time Series Generator pipeline.}
  \label{fig:chrongen}
  \vspace{-10pt} 
\end{wrapfigure}
To systematically construct multimodal datasets that align temporal dynamics with textual annotations, we propose \textbf{ChronGen}, a \emph{CHange-awaRe rules-OrieNtd time series GENerator}. As in \autoref{fig:chrongen}, ChronGen is designed to synthesize sequences that progressively incorporate diverse trends while maintaining explicit textual labels for each segment. This capability provides a controlled environment for studying multimodal reasoning, where both temporal evolution and natural language descriptions are essential.
Given a target length $L$ and the number of segments $K$, ChronGen first samples $K-1$ change points along the timeline, which divide the sequence into contiguous segments. For the initial segment, a base trend (e.g., linear, constant, or oscillatory) is instantiated. Each subsequent segment is generated by conditioning on the previous trend and injecting additional variations to gradually enrich the dynamics. This process yields a piecewise-evolving time series that captures both continuity and variability.
For every generated segment, ChronGen produces a corresponding textual annotation, which describes the dominant trend (e.g., ``increasing steadily,'' ``plateau rising,'' ``decreasing sharply''). These annotations serve as aligned labels, ensuring that each temporal unit has a semantic description that can be leveraged in downstream multimodal learning. As a result, ChronGen provides a flexible and interpretable way to generate synthetic data that balances temporal complexity with linguistic clarity.

\subsection{Model training}
\label{subsec:chrongen_training_datasets}
To endow models with the ability to reason about and analyze time series that exhibit rich and evolving dynamics, we construct large-scale training data using the proposed \textsc{ChronGen} framework. Specifically, ChronGen generates multivariate time series characterized by diverse change patterns and temporal complexities. Each sequence is paired with textual annotations that describe the segment-wise dynamics, thereby aligning temporal signals with natural-language semantics.

To further enhance the reasoning dimension, we design task-oriented question–answer templates, which transform the generated time series and annotations into training samples for multimodal reasoning. These templates encourage the model to connect raw temporal evolution with high-level textual inference, reflecting real-world analytical demands. The detailed configurations of the templates are provided in \autoref{subsec:training_qa_template}.

Through this process, our dataset integrates synthetic yet interpretable temporal trajectories with semantically grounded textual descriptions, offering a controlled but expressive environment for training multimodal time-series reasoning models.

During model training, we first adopted the time-series data generation strategy proposed in ~\cite{xie2024chatts} to construct an initial alignment dataset, comprising 100K question–answer pairs sampled from a mixture of the original datasets. This step was designed to endow the model with fundamental alignment capabilities. Building upon this foundation, we further performed supervised fine-tuning (SFT) on another 100K training instances generated using the same method, thereby enabling deeper enhancement of the model’s reasoning and task-solving abilities. To further strengthen the model’s instruction-following capacity, we additionally incorporated the Tulu instruction dataset from ~\cite{lambert2024tulu}. The full training configurations are detailed in ~\autoref{subsec:app_training_details}.

\subsection{EvalTS}
\label{subsec:tsbench}

\begin{figure}[t]
    \centering
    \includegraphics[width=0.95\linewidth]{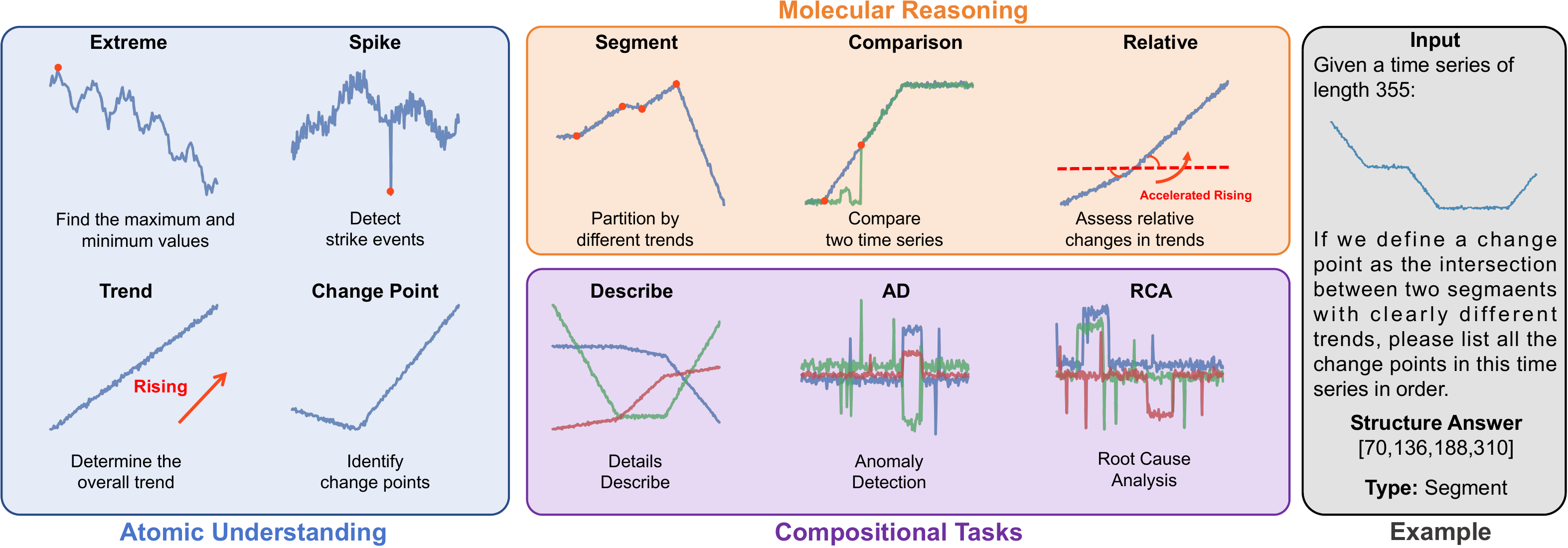}
    \caption{
        Overview of task categories in \tsbench. 
        }
    \label{fig:ts_bench}
    \vspace{-15pt}
\end{figure}

As shown in \autoref{fig:ts_bench}, the EvalTS benchmark decomposes multimodal time-series evaluation into three categories: Atomic Understanding, Molecular Reasoning, and Compositional Tasks. 
Illustrative examples are provided in \autoref{subsec:app_evalts_exampls}.

\textit{\textbf{Atomic Understanding.}}
This category targets the most fundamental units of time-series cognition: individual points and global trends. Tasks include extremum identification, overall trend recognition, change-point detection, strike detection, and value retrieval at a specific index. Tasks are divided into single-series and multi-series settings, and sequence length is unconstrained, reflecting real-world scenarios with variable-duration series. 

\textit{\textbf{Molecular Reasoning.}}
Building on atomic cognitions, this category requires establishing relationships between multiple fundamental units for dynamic temporal understanding. Tasks include segmenting a series into local trends, comparing trends across two series, and assessing relative changes between consecutive trends. These tasks demand integration of trend and value information to evaluate temporal patterns and relative dynamics.

\textit{\textbf{Compositional Tasks.}}
This category assesses the application of atomic and molecular reasoning in complex scenarios. Tasks include analyzing multivariate or complex series, detecting anomalies based on user-defined rules, and ranking anomalies while identifying root causes. These tasks require combining multiple basic abilities and are designed to closely reflect real-world temporal reasoning challenges.

\section{Experiments}
\label{sec:eval}
\begin{table}[t]
    \centering
    \caption{Model Performance on Fundamental and Specialized Tasks (Avg. Columns Removed)}
    \vspace{-10pt}
    \label{tab:all_tasks_no_avg}
    \adjustbox{width=\linewidth}{%
    \begin{tabular}{l|cccccccc|ccc|ccc}
        \toprule
        \multirow{3}{*}{\textbf{Model}} 
        & \multicolumn{8}{c|}{\textbf{Atomic Understanding}} 
        & \multicolumn{3}{c|}{\textbf{Molecular Reasoning}} 
        & \multicolumn{3}{c}{\textbf{Compositional Tasks}} \\
    \cmidrule(lr){2-9} \cmidrule(lr){10-12} \cmidrule(lr){13-15}
        & \multicolumn{2}{c}{Change Point} 
        & \multicolumn{2}{c}{Extreme} 
        & \multicolumn{2}{c}{Spike} 
        & \multicolumn{2}{c|}{Trend} 
        & \multirow{2}{*}{Segment} 
        & \multirow{2}{*}{Comparison} 
        & \multirow{2}{*}{Relative} 
        & \multirow{2}{*}{Describe} 
        & \multirow{2}{*}{RCA} 
        & \multirow{2}{*}{AD} \\
    \cmidrule(lr){2-3} \cmidrule(lr){4-5} \cmidrule(lr){6-7} \cmidrule(lr){8-9}
        & Uni & Multi & Uni & Multi & Uni & Multi & Uni & Multi & & & & & & \\
        \midrule
        ChatTime 
        & -- & -- & -- & -- & -- & -- & 0.85 & 0.31 
        & -- & -- & -- 
        & -- & -- & -- \\

        Time-MQA 
        & 0.06 & 0.19 & 0.56 & 0.27 & 0.21 & 0.24 & 0.41 & 0.37 
        & 0.13 & 0.14 & 0.02 
        & 0.05 & 0.01 & 0.02 \\                 

        Qwen2.5 
        & 0.18 & 0.05 & 0.86 & 0.56 & 0.30 & 0.37 & 0.95 & 0.68 
        & 0.15 & 0.13 & 0.58 
        & 0.05 & 0.01 & 0.00 \\

        ChatTS 
        & 0.03 & 0.06 & 0.30 & 0.01 & 0.79 & 0.56 & \textcolor{red}{\textbf{0.99}} & 0.77
        & 0.00 & 0.32 & 0.54
        & \textcolor{blue}{\textbf{0.14}} & \textcolor{blue}{\textbf{0.36}} & 0.46 \\

        GPT-5 
        & \textcolor{blue}{\textbf{0.32}} & \textcolor{blue}{\textbf{0.35}} & \textcolor{blue}{\textbf{0.98}} & \textcolor{blue}{\textbf{0.89}}
        & \textcolor{blue}{\textbf{0.90}} & \textcolor{blue}{\textbf{0.84}} & 0.95 & 0.71
        & \textcolor{blue}{\textbf{0.34}} & \textcolor{blue}{\textbf{0.50}} & \textcolor{blue}{\textbf{0.66}}
        & 0.13 & 0.23 & 0.46 \\
        \midrule

        TimeSenses-7B 
        & 0.05 & 0.15 & 0.93 & 0.44 & 0.23 & 0.24 & 0.98 & \textcolor{blue}{\textbf{0.94}}
        & 0.19 & 0.33 & 0.53
        & 0.10 & 0.33 & \textcolor{blue}{\textbf{0.49}} \\

        TimeSense-14B 
        & \textbf{\textcolor{red}{0.98}} & \textbf{\textcolor{red}{0.97}} & \textbf{\textcolor{red}{0.99}} & \textbf{\textcolor{red}{0.97}}
        & \textbf{\textcolor{red}{0.95}} & \textbf{\textcolor{red}{0.87}} & \textbf{\textcolor{red}{0.99}} & \textbf{\textcolor{red}{0.98}}
        & \textcolor{red}{\textbf{0.88}} & \textcolor{red}{\textbf{0.84}} & \textcolor{red}{\textbf{0.84}}
        & \textcolor{red}{\textbf{0.39}} & \textcolor{red}{\textbf{0.49}} & \textcolor{red}{\textbf{0.82}} \\

        \bottomrule
    \end{tabular}}
    \vspace{-10pt}
\end{table}

\subsection{Experiment Setup}
\textbf{Evaluation Datasets.} 
We evaluate all models on the EvalTS introduced in \autoref{subsec:tsbench}, where task definitions and example queries are described in \autoref{subsec:app_evalts_exampls}. 
To expand the size of the evaluation set, we choose an open-source dataset used by ChatTime \cite{wang2024chattime}.

We selected four types of multiple-choice questions that reflect temporal reasoning capabilities and categorized them into cross-domain and out-of-domain tasks. Specifically, two datasets focusing on temporal trends and outliers were used as cross-domain evaluation sets, denoted as MCQA D1 and MCQA D2. Datasets containing seasonality and volatility features, which are not present in our training data, were used as out-of-domain evaluation sets and displayed 
as MCQA D3 and MCQA D4.
\textbf{Evaluation Models.}
In particular, we train {TimeSense-14B} and compare against {GPT-5}, {ChatTS}~\citep{xie2024chatts}, and {Qwen2.5}~\citep{bai2025qwen2}. Specifically, Qwen2.5 refers to Qwen2.5-14B-Instruct, while ChatTS is based on this model as its backbone. 
For parameter-scale comparison, we also train and evaluate {TimeSense-7B} alongside {Time-MQA}~\citep{kong2025time} and {ChatTime}~\citep{wang2024chattime}, where both {ChatTime} and {Time-MQA} are 7B-scale models. 
Since {ChatTime} only supports multiple-choice outputs, we adapt the evaluation format for the \textit{Trend} task into a multiple-choice style to make the evaluation feasible.

\subsection{Experiment Results}
\begin{table*}[t]
\centering
\scriptsize
\caption{Model performance across cross-domain (CD) and out-of-domain (OOD) tasks. }
\vspace{-10pt}
\label{tab:exp_all_mcqa}
\begin{adjustbox}{width=\linewidth}
\begin{tabular}{l|cccc|cccc|cccc|cccc}
\toprule
\textbf{Model} & \multicolumn{4}{c|}{\textbf{CD D1}} & \multicolumn{4}{c|}{\textbf{CD D2}} & \multicolumn{4}{c|}{\textbf{OOD D3}} & \multicolumn{4}{c}{\textbf{OOD D4}} \\
\cmidrule(lr){2-5} \cmidrule(lr){6-9} \cmidrule(lr){10-13} \cmidrule(lr){14-17}
& 64 & 128 & 256 & 512 & 64 & 128 & 256 & 512 & 64 & 128 & 256 & 512 & 64 & 128 & 256 & 512 \\
\midrule
ChatTime & \textbf{\textcolor{blue}{0.90}} & \textbf{\textcolor{blue}{0.90}} & 0.88 & 0.82 
         & 0.79 & 0.70 & 0.62 & 0.57
         & \textbf{\textcolor{blue}{0.66}} & \textbf{\textcolor{red}{0.65}} & \textbf{\textcolor{blue}{0.65}} & \textbf{\textcolor{blue}{0.62}}
         & \textbf{\textcolor{red}{0.88}} & \textbf{\textcolor{red}{0.90}} & \textbf{\textcolor{red}{0.86}} & \textbf{\textcolor{red}{0.75}} \\
Time-MQA & 0.41 & 0.42 & 0.32 & 0.41 & 0.33 & 0.50 & 0.27 & 0.35
         & \textbf{\textcolor{red}{0.73}} & 0.52 & 0.40 & 0.42
         & 0.45 & 0.35 & 0.18 & 0.35 \\
Qwen2.5 & 0.25 & 0.36 & 0.21 & 0.48 & 0.32 & 0.36 & 0.32 & 0.28
        & 0.54 & 0.62 & 0.52 & 0.44
        & 0.34 & 0.50 & 0.46 & 0.47 \\
ChatTS & 0.54 & 0.88 & \textbf{\textcolor{blue}{0.91}} & \textbf{\textcolor{blue}{0.92}}
       & 0.67 & \textbf{\textcolor{blue}{0.96}} & \textbf{\textcolor{red}{0.99}} & \textbf{\textcolor{red}{0.99}}
       & 0.33 & 0.38 & 0.36 & 0.25
       & 0.39 & 0.61 & 0.61 & 0.61 \\
GPT5 & 0.81 & 0.54 & 0.46 & 0.67 & \textbf{\textcolor{blue}{0.84}} & 0.65 & 0.79 & 0.68
     & 0.63 & 0.50 & 0.63 & 0.54
     & 0.42 & 0.50 & 0.32 & 0.43 \\
\midrule
TimeSense-7B & 0.86 & 0.85 & 0.85 & 0.86 & 0.72 & 0.78 & 0.72 & 0.73
             & 0.18 & 0.27 & 0.25 & 0.27
             & 0.25 & 0.32 & 0.26 & 0.17 \\
TimeSense-14B & \textbf{\textcolor{red}{0.94}} & \textbf{\textcolor{red}{0.92}} & \textbf{\textcolor{red}{0.95}} & \textbf{\textcolor{red}{0.93}}
              & \textbf{\textcolor{red}{0.99}} & \textbf{\textcolor{red}{0.98}} & \textbf{\textcolor{red}{0.99}} & \textbf{\textcolor{red}{0.99}}
              & 0.62 & \textbf{\textcolor{red}{0.65}} & \textbf{\textcolor{red}{0.67}} & \textbf{\textcolor{red}{0.65}}
              & \textbf{\textcolor{blue}{0.72}} & \textbf{\textcolor{blue}{0.68}} & \textbf{\textcolor{blue}{0.67}} & \textbf{\textcolor{blue}{0.69}} \\
\bottomrule
\end{tabular}
\end{adjustbox}
\vspace{-15pt}
\end{table*}

The EvalTS results are summarized in \autoref{tab:all_tasks_no_avg}. Across nearly all categories, {TimeSense-14B} achieves the best performance, substantially outperforming other baselines. This demonstrates the effectiveness of explicitly incorporating temporal reasoning modules into multimodal LLMs.
We make the following observations:

\textbf{Comparison to LLM.}  
Mainstream text-based models such as {GPT-5} show strong results on fundamental tasks (e.g., Extreme, Index, and Spike), likely due to their superior language understanding and pretraining scale. However, their advantage diminishes in compound and complex temporal reasoning tasks, where temporal dynamics are more critical.
Compared with text-only {Qwen-14B}, both {ChatTS-14B} and {TimeSense-14B} consistently perform better, especially on complex tasks (e.g., RCA, AD). Interestingly, on several datasets, they even surpass {GPT-5}, highlighting that multimodal temporal modules better capture sequential dependencies for reasoning.

\textbf{Comparison to TS-MLLM.}  
When scaling down to 7B parameters, {TimeSense-7B} still surpasses {Time-MQA-7B} and {ChatTime-7B}. This suggests that the design of {TimeSense} provides stronger generalization, though smaller models naturally show degraded performance compared to their 14B counterparts.

Overall, these results confirm that {TimeSense} not only scales effectively but also consistently outperforms both text-only and multimodal baselines across different task categories, especially in complex temporal reasoning.

\subsection{Generalization of TimeSense}
As exhibited in \autoref{tab:exp_all_mcqa} , this evaluation consists of two parts, namely Cross-Domain tasks and Out-of-Domain tasks, depending on their relevance to TimeSense’s training data. TimeSense-14B achieves consistent superiority on Cross-Domain tasks and remains competitive with ChatTime on Out-of-Domain tasks, despite the latter being trained with domain data. When scaling down to TimeSense-7B, the model performs on par with ChatTime in Cross-Domain tasks and even shows advantages on longer series. However, it exhibits a notable performance drop in Out-of-Domain scenarios, highlighting the dependency of generalization ability on model size.

\subsection{Ablation Studies}
\label{subsec:ablation}

\begin{figure*}[t]
  \centering
  \begin{subfigure}[b]{0.32\linewidth}
    \centering
    \includegraphics[width=\linewidth]{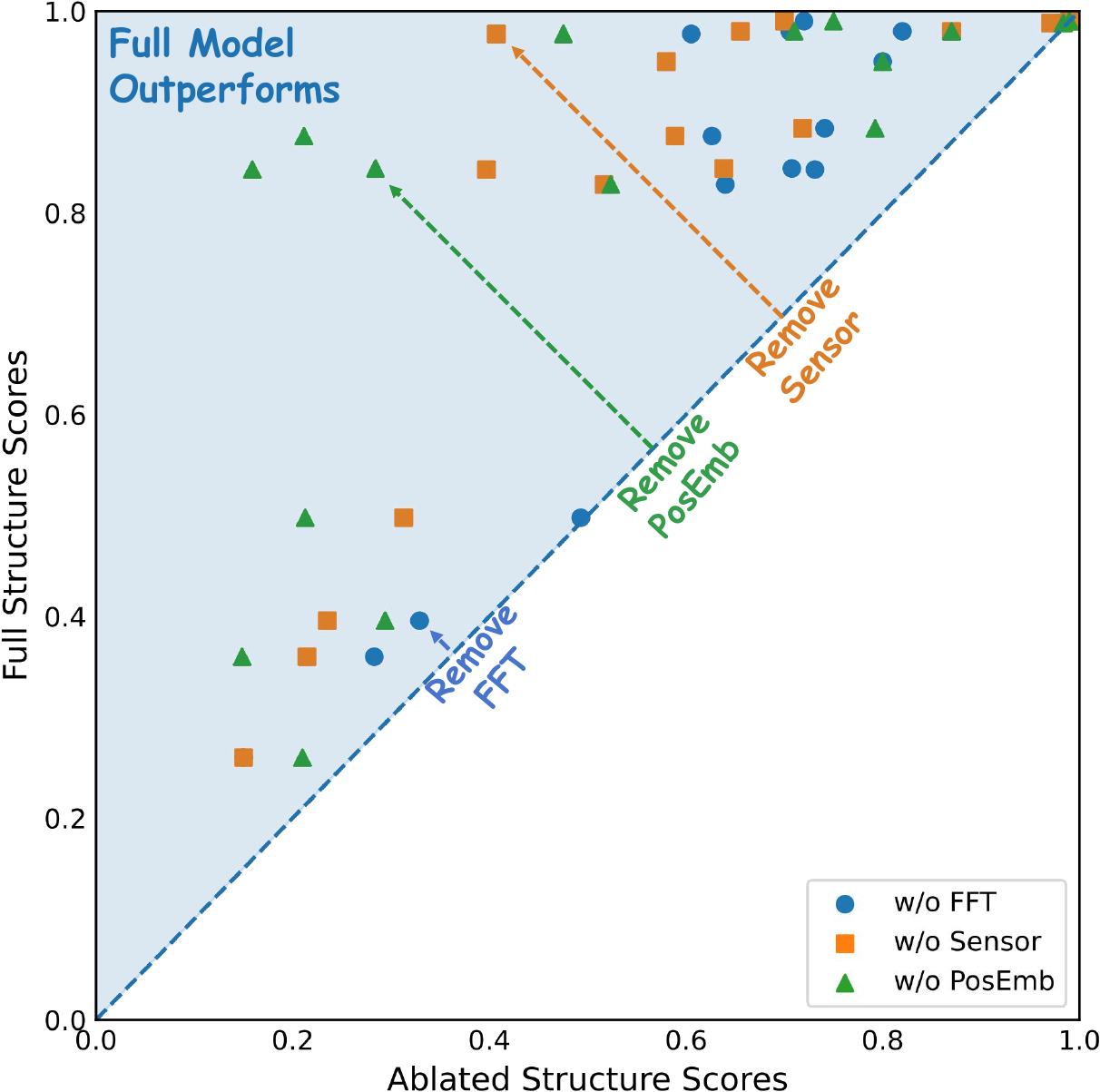}
    \caption{Comparison across all tasks}
    \label{fig:aba_structure2}
  \end{subfigure}
  \hfill
  \begin{subfigure}[b]{0.55\linewidth} 
    \centering
    \includegraphics[width=\linewidth]{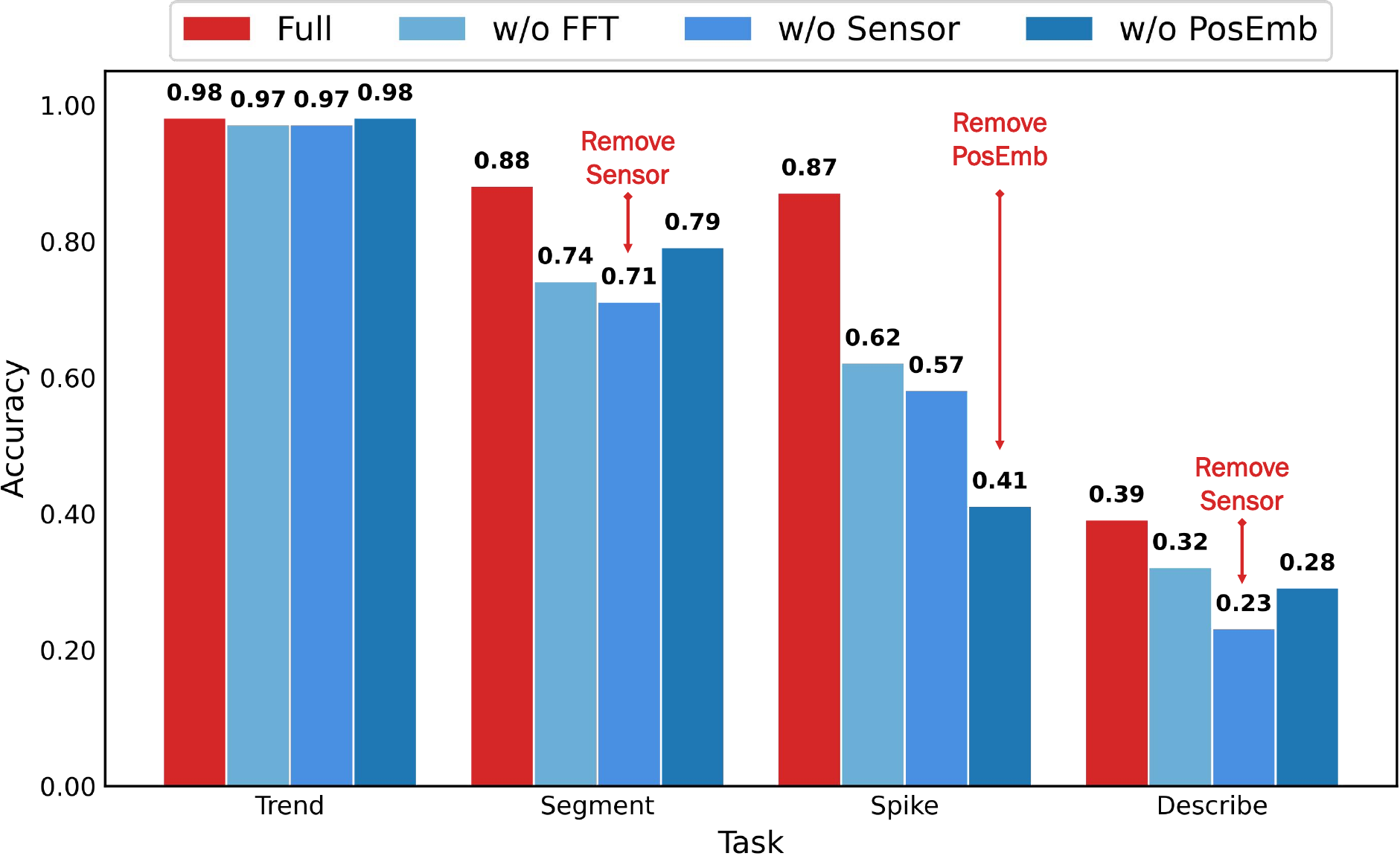}
    \caption{Comparison across representative tasks}
    \label{fig:aba_structure}
  \end{subfigure}
  \caption{Ablation studies of different model structures}
  \vspace{-15pt}  
  \label{fig:aba_combined}
\end{figure*}

\textbf{Effectiveness of Different Model Structures.}
To evaluate the contributions of time-series position embedding (PosEmb), sensor(Sensor), and FFT loss(FFT) in training, we conduct ablation studies by removing each component. The results are shown in Figure~\ref{fig:aba_structure2}. Detailed results are provided in Section~\ref{subsec:ablation_study_details}. For the basic Trend task, all variants remain strong, showing robustness to minor changes. However, Segment exhibits sharp declines, especially when removing Sensor and FFT loss (up to 15\%), highlighting the Sensor’s role in capturing global temporal shifts and the FFT loss in refining periodic sensitivity. For Spike task, ablating Time-Series PosEmb causes the largest drop, confirming its importance for precise temporal localization. Finally, Describe shows broader degradation under Sensor ablation, reflecting its reliance on multimodal integration.

\section{Conclusion}
\label{sec:conclusion}

In this work, we presented TimeSense, a multimodal time series model that integrates a Temporal Sense module to balance textual and temporal information, addressing the bias toward language in prior approaches. To characterize complex temporal reasoning, we introduced ChronGen, a controllable generator for multidimensional time series with trend variations, anomalies, and cross-channel interactions, and built the EvalTS benchmark spanning ten tasks across three difficulty levels. Experiments demonstrate that TimeSense achieves state-of-the-art performance. 

\newpage


\bibliography{iclr2026_conference}
\bibliographystyle{iclr2026_conference}
\newpage
\appendix
\section{appendix}
\label{sec:appendix}
You may include other additional sections here.
\subsection{training qa template}
\label{subsec:training_qa_template}
In this section, we provide illustrative examples of the data generation procedure described in \autoref{subsec:chrongen_training_datasets}. The Question and Answer components are used for SFT training, where the temporal information is extracted into a dedicated time-series modality that serves both as input and as the reconstruction target. The Feature component highlights the internal design of ChronGen, including the step-by-step generation of change points, the trend patterns associated with the segments defined by these change points, and optional spike values that capture abrupt local variations.
\begin{figure}[ht]
    \centering
    \includegraphics[width=0.9\linewidth]{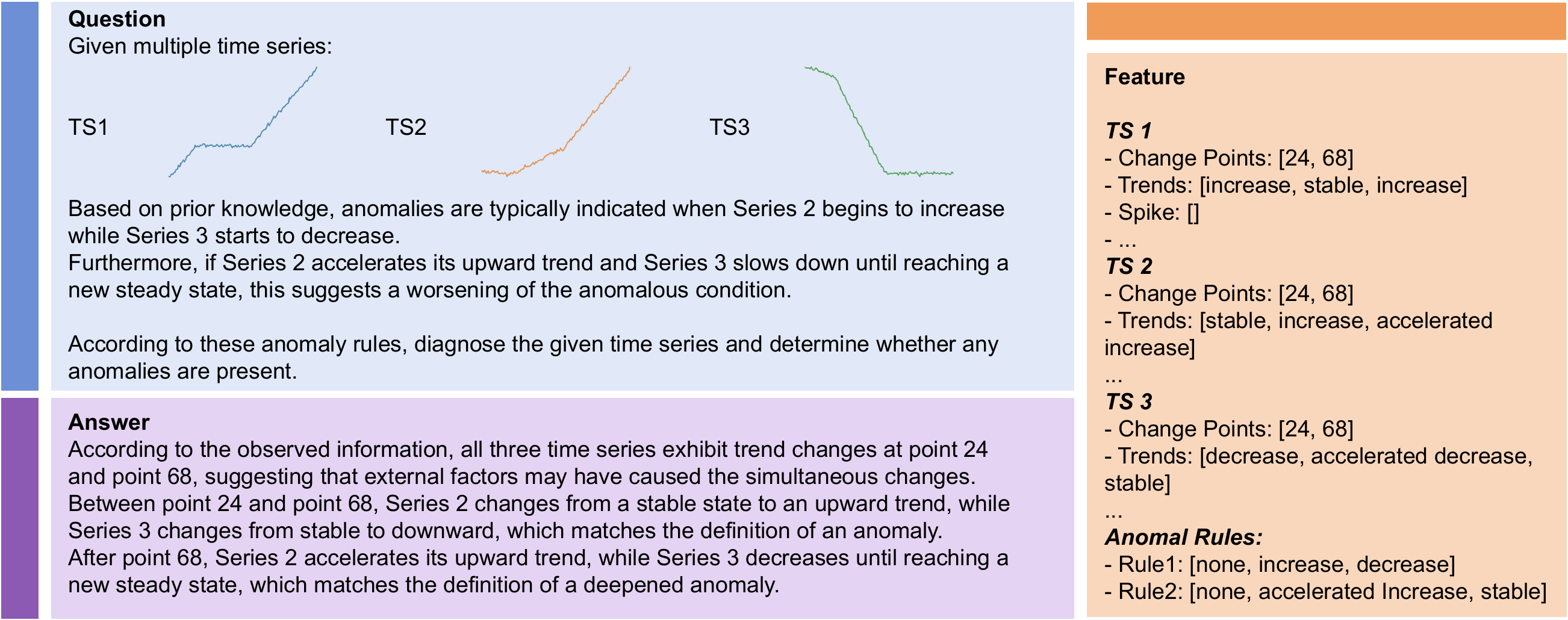}
    \caption{Examples of training datasets}
    \label{fig:app_chrongen_example}
\end{figure}

\subsection{chrongen code}
\label{subsec:chrongen}
The generator produces a piecewise time series by first splitting the timeline into $K$ segments. Each segment starts with a base trend $T_i$, and richer dynamics are introduced by iteratively applying change functions (e.g., \texttt{point\_change}, \texttt{segment\_change}). Each change is immediately recorded in a feature array $\mathbf{F}$ and simultaneously converted into a textual annotation via a text template function. This ensures that every incremental modification to the time series has a corresponding annotation, providing fine-grained supervision for both numerical trends and textual descriptions.
\begin{algorithm}[ht]
\caption{Change-oriented Time Series Generator with Incremental Annotation}
\label{alg:tscgen}
\begin{algorithmic}[1]
\REQUIRE time length $L$, number of segments $K$
\ENSURE time series $\mathbf{X}$ with text annotations $\mathbf{Y}$ and feature array $\mathbf{F}$
\STATE Initialize empty sequence $\mathbf{X} \gets []$, feature array $\mathbf{F} \gets []$, annotation set $\mathbf{Y} \gets []$
\STATE Randomly sample $K-1$ change points within $[1, L]$
\STATE Split timeline into segments $\{S_1, S_2, \dots, S_K\}$
\FOR{$i = 1$ to $K$}
    \STATE Generate base trend $T_i$
    \STATE Initialize segment sequence $S_i \gets []$
    \STATE Initialize segment annotation $Y_i \gets []$
    \STATE Iteratively construct variation $\Delta T_i$:
    \WHILE{more changes to apply}
        \STATE Sample a change operation $\text{change} \in \{\text{point\_change}, \text{segment\_change}, \dots\}$
        \STATE Update segment: $S_i \gets S_i + \text{change}(S_i, T_{i-1})$
        \STATE Record feature: $\mathbf{F} \gets \mathbf{F} \cup \{\text{change}\}$
        \STATE Generate textual annotation: $y \gets \text{text\_template}(\text{change})$
        \STATE Append $y$ to $Y_i$
    \ENDWHILE
    \STATE Append segment $S_i$ to $\mathbf{X}$
    \STATE Append segment annotations $Y_i$ to $\mathbf{Y}$
\ENDFOR
\STATE \textbf{return} $\mathbf{X}$, $\mathbf{Y}$, and feature array $\mathbf{F}$
\end{algorithmic}
\end{algorithm}

\subsection{Training Details}
\label{subsec:app_training_details}

In the initial stage of time-series--text alignment, we integrated the Stage 1 and Stage 2 training procedure of~\cite{xie2024chatts}. Specifically, their released datasets include four instruction-tuning corpora: \textsc{ChatTS-IFT} for instruction following, \textsc{ChatTS-SFT} for question answering, and two alignment-focused datasets, \textsc{ChatTS-Align\_256} and \textsc{ChatTS-Align\_random}. Different from~\cite{xie2024chatts}, we constructed a mixed dataset by combining \textsc{ChatTS-Align\_random}, \textsc{ChatTS-SFT}, and \textsc{ChatTS-IFT} with a ratio of $5{:}3{:}2$, and randomly sampled 100K pairs as the training corpus. This enabled us to accomplish model alignment training within a single stage.

In the second stage, we adopted the data generators introduced in~\autoref{subsec:chrongen_training_datasets} to produce 100K time-series–text pairs. These raw pairs were then converted into task-specific question–answer formats through a set of designed QA rules, resulting in the \textsc{TS-Enhance} dataset. To further equip the model with capabilities for domain-specific tasks, we additionally incorporated a portion of time-series reasoning data derived from proprietary corpora, enhancing the model’s ability to handle practical, real-world scenarios.

Both stages employed identical training hyperparameters. We performed full-parameter supervised fine-tuning (SFT) using \textsc{DeepSpeed} and \textsc{LlamaFactory}~\citep{zheng2024llamafactory} on a single machine with 8$\times$A800 GPUs. The detailed hyperparameters are summarized in \autoref{tab:training_hyperparameters}.

\begin{table}[ht]
\centering
\caption{Training hyperparameters used in both Stage 1 and Stage 2.}
\label{tab:training_hyperparameters}
\begin{tabular}{l|c}
\toprule
\textbf{Hyperparameter} & \textbf{Value} \\
\midrule
Per-device train batch size & 1 \\
Gradient accumulation steps & 32 \\
Maximum training steps & 1200 \\
Learning rate & 1e-5 \\
Warmup ratio & 0.02 \\
GPUs & 8 $\times$ A800 \\
\bottomrule
\end{tabular}
\end{table}

\subsection{EvalTS examples}
\label{subsec:app_evalts_exampls}
In this section, we present examples for the majority of the tasks. The question represents the multimodal input directly fed into the model, while the answer feature serves as the reference for evaluation. We first extract the corresponding predicted feature from the model’s output using a large language model (LLM), and then perform a one-to-one comparison between the predicted and reference features to compute the evaluation accuracy.
\begin{figure}
    \centering
    \includegraphics[width=\linewidth]{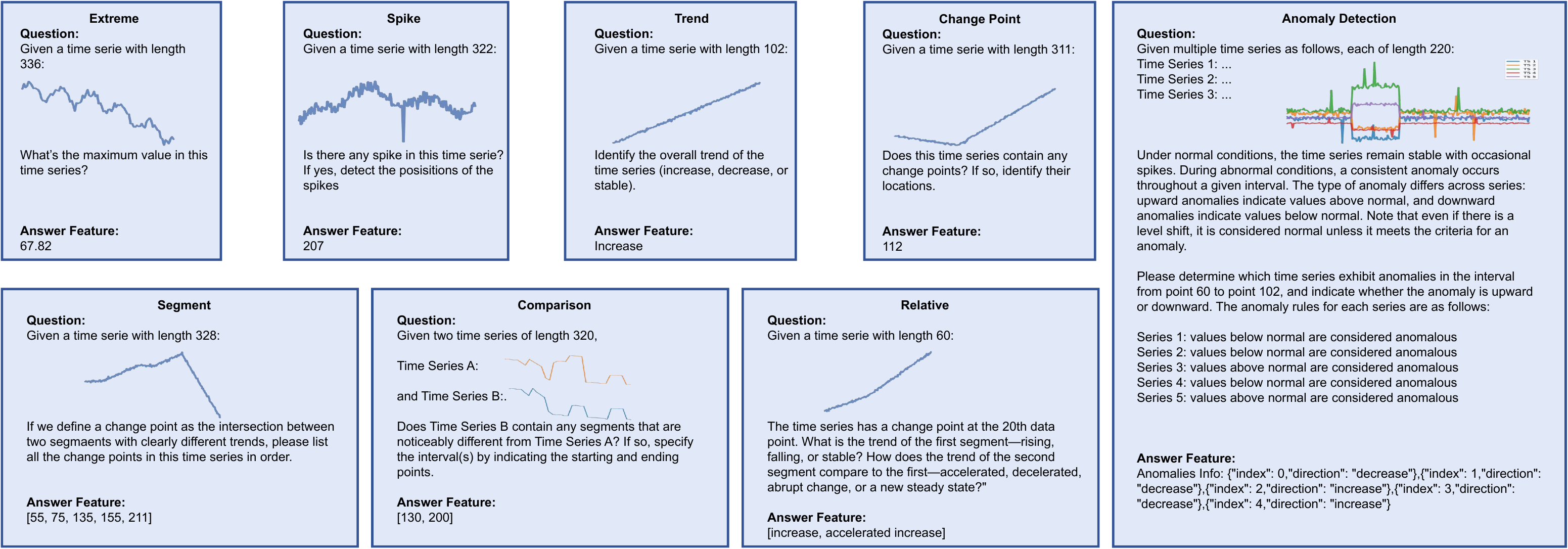}
    \caption{EvalTS examples}
    \label{fig:app_tseval_examples}
\end{figure}

\subsection{ablation study details}
\label{subsec:ablation_study_details}
Since we present the complete results of the ablation studies on both the model architecture and patch size. Overall, the findings demonstrate that each component contributes distinctly to the model’s capabilities: time series PosEmb modules enhance temporal precision, time series sensor provide global contextual awareness, and certain loss functions improve pattern modeling. The effects of patch size are generally task-dependent, reflecting a trade-off between capturing broader temporal context and maintaining fine-grained resolution. Across the board, the results highlight how architectural design choices and temporal representation granularity jointly shape the model’s overall performance and robustness.
\begin{table}[ht]
    \centering
    \caption{Model Performance on different patch size}
    \label{tab:ablations_full}
    \adjustbox{width=\linewidth}{%
    \begin{tabular}{l|cccccccc|ccc|ccc}
    \toprule
    \multirow{3}{*}{\textbf{Model}} 
        & \multicolumn{8}{c|}{\textbf{Fundamental Tasks}} 
        & \multicolumn{3}{c|}{\textbf{Compound Tasks}} 
        & \multicolumn{3}{c}{\textbf{Complex Tasks}} \\
    \cmidrule(lr){2-9} \cmidrule(lr){10-12} \cmidrule(lr){13-15}
        & \multicolumn{2}{c}{Change Point} 
        & \multicolumn{2}{c}{Extreme} 
        & \multicolumn{2}{c}{Spike} 
        & \multicolumn{2}{c|}{Trend} 
        & \multirow{2}{*}{Segment} 
        & \multirow{2}{*}{Comparison} 
        & \multirow{2}{*}{Relative} 
        & \multirow{2}{*}{Describe} 
        & \multirow{2}{*}{RCA} 
        & \multirow{2}{*}{AD} \\
    \cmidrule(lr){2-3} \cmidrule(lr){4-5} \cmidrule(lr){6-7} \cmidrule(lr){8-9}
        & Uni & Multi & Uni & Multi & Uni & Multi & Uni & Multi & & & & & & \\
    \midrule
    4padded  & 0.99 & 0.98 & 0.99 & 0.97 & 0.97 & 0.89 & 0.99 & 0.97 & 0.80 & 0.84 & 0.82 & 0.11 & 0.43 & 0.81 \\
    8padded  & 0.98 & 0.97 & 0.99 & 0.97 & 0.95 & 0.87 & 0.99 & 0.98 & 0.88 & 0.84 & 0.84 & 0.39 & 0.49 & 0.82 \\
    16padded & 0.83 & 0.76 & 0.99 & 0.96 & 0.87 & 0.83 & 0.99 & 0.99 & 0.87 & 0.76 & 0.89 & 0.24 & 0.49 & 0.82 \\
    32padded & 0.31 & 0.21 & 0.99 & 0.96 & 0.87 & 0.78 & 0.99 & 0.97 & 0.60 & 0.42 & 0.36 & 0.03 & 0.48 & 0.83 \\
    64padded & 0.22 & 0.19 & 0.99 & 0.90 & 0.72 & 0.29 & 0.99 & 0.97 & 0.14 & 0.35 & 0.18 & 0.04 & 0.47 & 0.81 \\
    \bottomrule
    \end{tabular}}
\end{table}

\begin{table}[ht]
\centering
\caption{Ablation Study on Benchmark Tasks}
\label{tab:ablation_tasks_no_avg}
\adjustbox{width=\linewidth}{%
\begin{tabular}{l|cccccccc|ccc|ccc}
\toprule
\multirow{3}{*}{\textbf{Model}} 
    & \multicolumn{8}{c|}{\textbf{Fundamental Tasks}} 
    & \multicolumn{3}{c|}{\textbf{Compound Tasks}} 
    & \multicolumn{3}{c}{\textbf{Complex Tasks}} \\
\cmidrule(lr){2-9} \cmidrule(lr){10-12} \cmidrule(lr){13-15}
    & \multicolumn{2}{c}{Change Point} 
    & \multicolumn{2}{c}{Extreme} 
    & \multicolumn{2}{c}{Spike} 
    & \multicolumn{2}{c|}{Trend} 
    & \multirow{2}{*}{Segment} 
    & \multirow{2}{*}{Comparison} 
    & \multirow{2}{*}{Relative} 
    & \multirow{2}{*}{Describe} 
    & \multirow{2}{*}{RCA} 
    & \multirow{2}{*}{AD} \\
\cmidrule(lr){2-3} \cmidrule(lr){4-5} \cmidrule(lr){6-7} \cmidrule(lr){8-9}
    & Uni & Multi & Uni & Multi & Uni & Multi & Uni & Multi & & & & & & \\
\midrule
Full          & 0.98 & 0.97 & 0.99 & 0.97 & 0.95 & 0.87 & 0.99 & 0.98 & 0.88 & 0.84 & 0.84 & 0.39 & 0.49 & 0.82 \\
w/o-FFT       & 0.82 & 0.60 & 0.98 & 0.97 & 0.80 & 0.62 & 0.99 & 0.97 & 0.74 & 0.70 & 0.73 & 0.32 & 0.49 & 0.63 \\
w/o-Sensor    & 0.87 & 0.40 & 0.97 & 0.97 & 0.58 & 0.58 & 0.99 & 0.97 & 0.71 & 0.63 & 0.39 & 0.23 & 0.31 & 0.51 \\
w/o-PosEmb    & 0.87 & 0.47 & 0.99 & 0.96 & 0.80 & 0.41 & 0.99 & 0.98 & 0.79 & 0.28 & 0.15 & 0.29 & 0.21 & 0.52 \\
\bottomrule
\end{tabular}}
\end{table}

\end{document}